\documentclass{article}

\usepackage[preprint,nonatbib]{neurips_2026}
\usepackage[utf8]{inputenc} 
\usepackage[backend=biber,style=numeric]{biblatex}

\usepackage[T1]{fontenc}    
\usepackage{hyperref}       
\usepackage{url}            
\usepackage{booktabs}       
\usepackage{amsfonts}       
\usepackage{nicefrac}       
\usepackage{microtype}      
\usepackage{xcolor}         
\usepackage{color}
\usepackage{soul}
\usepackage{amsmath}
\usepackage{xspace}
\usepackage{multirow}
\usepackage{enumitem}
\usepackage{caption}
\usepackage{graphicx}
\usepackage{wrapfig}
\usepackage[inkscapepath=svgcache]{svg} 
\usepackage{algorithm}
\usepackage{algorithmic}
\usepackage{float}
\usepackage{makecell}

\newcommand{\R}{\mathbb{R}}
\newcommand{\N}{\mathbb{N}}
\newcommand{\method}{\textsc{MatryoshkaLora}\xspace}

\newcommand{\gray}[1]{\textcolor{gray}{#1}}
\newcommand{\bs}[1]{\ensuremath{\boldsymbol{\textcolor{blue}{#1}}}}

\title{\method: Learning Accurate Hierarchical Low-Rank Representations for LLM Fine-Tuning}
\author{
	Ionut-Vlad Modoranu\thanks{Institute of Science and Technology Austria (ISTA). Correspondence to \texttt{ionut-vlad.modoranu@ista.ac.at}} \\
	ISTA
	\And
	Mher Safaryan\\
	Lancaster University, UK
	\And
	Dan Alistarh\\
	ISTA
}

\begin{document}

\maketitle

\begin{abstract}
With the rise in scale for deep learning models to billions of parameters, the computational cost of fine-tuning remains a significant barrier to deployment. While Low-Rank Adaptation (LoRA) has become the standard for parameter-efficient fine-tuning, the need to set a predefined, static rank $r$ requires exhaustive grid searches to balance efficiency and performance. Existing rank-adaptive solutions such as DyLoRA mitigate this by sampling ranks during the training from a predefined distribution. However, they often yield sub-optimal results at higher ranks due to lack of consistent gradient signals across the full hierarchy of ranks, thus making these methods data-inefficient. In this paper, we propose \textsc{MatryoshkaLoRA}, a general, Matryoshka-inspired training framework for LoRA that learns accurate hierarchical low-rank representations by inserting a fixed, carefully crafted diagonal matrix $P$ between the existing LoRA adapters to scale their sub-ranks accordingly. By introducing this simple modification, our general framework recovers LoRA and DyLoRA only by changing $P$ and ensures all sub-ranks embed the available gradient information efficiently. Our \textsc{MatryoshkaLoRA} supports dynamic rank selection with minimal degradation in accuracy. We further propose Area Under the Rank Accuracy Curve (AURAC), a metric that consistently evaluates the performance of hierarchical low-rank adapters. Our results demonstrate that \textsc{MatryoshkaLoRA} learns more accurate hierarchical low-rank representations than prior rank-adaptive approaches and achieves superior accuracy-performance trade-offs across ranks on the evaluated datasets. Our code is available at \href{https://github.com/IST-DASLab/MatryoshkaLoRA}{\texttt{https://github.com/IST-DASLab/MatryoshkaLoRA}}.
\end{abstract}

\vspace{-1.5em}
\section{Introduction \& Related Work}\label{section:intro-related-work}
\begin{wrapfigure}{r}{0.4\textwidth}
  \centering
  \vspace{-1em}
  \includegraphics[width=0.4\textwidth]{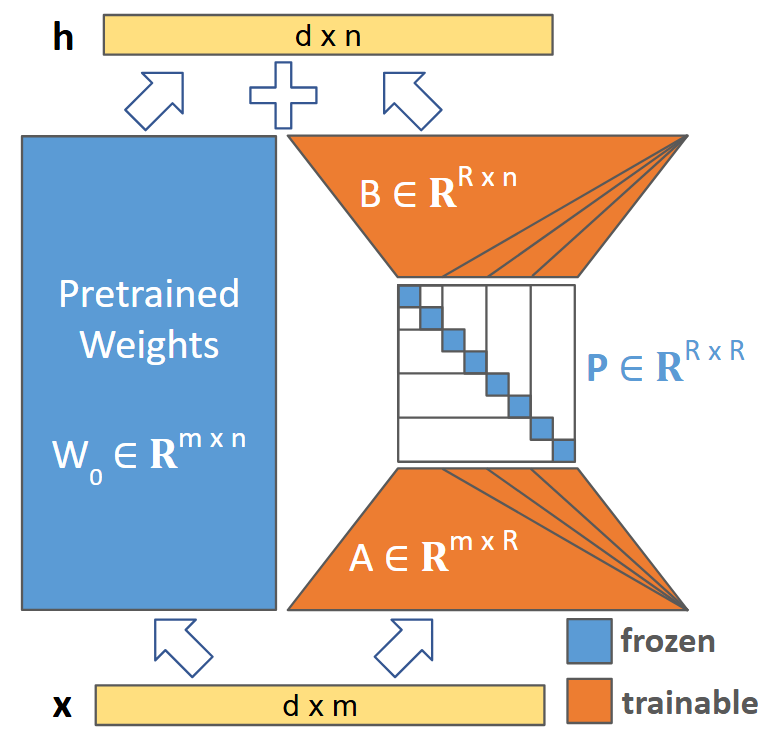}
  \caption{\method} \label{fig:schematic}
  \vspace{-1em}
\end{wrapfigure}
The research advancements in the Deep Learning community allowed researchers and practitioners to train models with billions of parameters. Building and managing pipelines for such a complex system requires a significant engineering effort, from preparing the dataset to scaling the model and the optimizer states via multi-dimensional parallelization techniques across many GPUs in a cluster or even multiple physical clusters - a process spanning over months. Given the costs, these large models are rarely deployed to production-ready environments as they are; instead, they serve as a knowledge base for downstream adaptation, such as fine-tuning, which still remains computationally prohibitive for many applications and settings.

To alleviate this overhead, LoRA~\cite{LoRA} has emerged as the de-facto standard for parameter-efficient fine-tuning (PEFT). However, it introduces a significant structural constraint: the rank $r$ must be predefined. Finding the optimal balance between parameter efficiency and model performance currently requires exhaustive search across multiple training runs, each with a different static rank. 

One line of work studies adaptive-rank LoRA methods, which optimize how rank capacity is allocated across layers or modules under a given or learned parameter budget. These methods improve parameter efficiency, but they typically produce a single specialized adapter configuration rather than a nested adapter family that can be sliced at inference time \cite{zhang2023adaloraadaptivebudgetallocation, ding2023sparselowrankadaptationpretrained, liu2024aloraallocatinglowrankadaptation, zhang2024autoloraautomaticallytuningmatrix, zhang2023increloraincrementalparameterallocation, singh2025l1radynamicrankassignment, cui2026iguloraadaptiverankallocation, kumaravelu2026postoptimizationadaptiverankallocation}.

A second direction studies dynamic-rank LoRA methods, which aim to train a single adapter whose prefixes are independently usable as lower-rank adapters. Unlike adaptive allocation methods, these approaches are designed to provide multiple sub-adapters from one checkpoint \cite{DyLoRA, LUO2025128778}. A conceptually related, but not directly comparable, line of work is slimmable \cite{yu2018slimmableneuralnetworks} and once-for-all networks \cite{cai2020onceforalltrainnetworkspecialize}. These are not LoRA methods, but they provide the broader once-for-all training paradigm: a single set of weights is trained so that many nested sub-networks are valid deployment choices \cite{yu2019universallyslimmablenetworksimproved}.

Our work can be interpreted as translating this paradigm from channel width to LoRA rank and belongs primarily to the second cluster. It differs from adaptive-rank allocation methods because it does not merely identify one rank configuration; it trains a nested adapter family. It differs from general slimmable networks because the nested structure is imposed on low-rank adapter factors rather than full-model channels. Its closest comparison is DyLoRA~\cite{DyLoRA}, which also targets dynamic rank usage but enforces rank hierarchy through a different training mechanism discussed next.

DyLoRA~\cite{DyLoRA} is an alternative solution to grid-searching for the rank $r$ for LoRA. At each training step, a rank $k \in \N$ is randomly sampled from a predefined distribution and the forward pass is performed using only the first $k$ rows and first $k$ columns of adapters $A$ and $B$, respectively, denoted by $A_k$ and $B_k$. The loss is computed with respect to the output of the network that was generated using only $A_k$ and $B_k$ for all linear layers. The purpose of this approach is to train adapters $A$ and $B$ that still yield high accuracy when using only a slice of size $k$ on the fly. DyLoRA solves the issue of adaptive rank $r$ only partially because, as we will show in our experimental section, the accuracies of the models fine-tuned with DyLoRA do not exhibit a true hierarchical pattern in the context of reasoning tasks, such as math datasets. We empirically show that DyLoRA strategy is sub-optimal because it constrains the learning only to a fixed rank $k$ sampled at random, while the ranks $r > k$ do not receive any gradient signal.

Our work proposes Matryoshka-style training framework for LoRA, inspired from Matryoshka Representation Learning~\cite{MatryoshkaRepresentationLearning}, which we call \method. It is motivated by the drawback of DyLoRA, which fails to learn what we call hierarchical low-rank features. Instead of using only one randomly generated rank $k$ per forward pass, we use all possible slices $A_r$ and $B_r$, with $r \in \{1, 2, 4, ..., R\}$, where $R$ is the maximum rank of $A$ and $B$. This way, the lower-dimensional representations are contained in higher-dimensional representations as prefixes, thus building a nested hierarchy of features. We present the schematic of \method in Figure~\ref{fig:schematic}.

We see three advantages of having an accurate technique to train hierarchical low-rank adapters: (1) it eliminates the need for multiple training runs to grid-search for different ranks, implicitly translating to a reduction in costs; (2) we benefit from a high-performance adapter at each rank $r$ by deploying targeted ranks on different devices depending on their computational power and (3) we enable dynamic rank selection under varying cluster loads to serve the requests at the same rate with minimal accuracy drop.

\textbf{Contribution.} We summarize our contribution as follows:
\vspace{-1em}
\begin{itemize}[leftmargin=*]
    \item We introduce a general framework inspired by Matryoshka Representation Learning for training LoRA adapters whose lower-rank prefixes remain accurate and independently usable at inference time, which views LoRA, DyLoRA, and Matryoshka-style adapters as different parameterizations of a shared rank-weighting vector;
    \item Starting from the goal of learning hierarchical low-rank representations at every layer, we show that this naturally leads to a simple diagonal weighting between the standard LoRA adapters $A$ and $B$, making the hierarchy explicit while keeping the implementation and exposition close to standard LoRA;
    \item We propose \textbf{\method}, a concrete instance of this framework that learns accurate nested low-rank prefixes via the new diagonal matrix. The resulting adapter can be evaluated or deployed at multiple ranks from a single checkpoint, without retraining separate adapters;
    \item We introduce \textbf{Area Under the Rank Accuracy Curve (AURAC)}, a metric for evaluating adapter performance across a set of ranks. AURAC summarizes the rank-performance tradeoff while weighting each rank according to its magnitude, reflecting the expectation that larger ranks should generally achieve stronger performance.
\end{itemize}
\vspace{-1em}
\section{Method}\label{section:method}
\vspace{-0.5em}
In this section, we introduce the notation we will use throughout the paper and provide details about our main baselines in the literature. Then we describe \method, our approach to learn accurate hierarchical low-rank representations for fine-tuning and AURAC, the metric we propose to assess the performance of LoRA approaches when evaluated on multiple ranks.
\subsection{Notation}
We consider the weights of a fully connected layer $W_0 \in \R^{m \times n}$ and the LoRA adapters $A \in \R^{m \times R}$ and $B \in \R^{R \times n}$, where $R$ is the maximum possible rank of $A$ and $B$ (bottleneck dimension), which is usually a power of $2$. Given the adapters $A$ and $B$ and an integer $k$, we extract the subsets of these adapters, denoted by $A_k = A[*, 1:k]$ (first $k$ columns of $A$) and $B_k = B[1:k, *]$ (first $k$ rows of $B$) by indexing into the $R$-dimension of each adapter.

We denote by $S$ the set of ranks used for training/inference. If not specified otherwise, the set $S$ is the same for both training and inference. Note that in our work we restrict $S$ to contain only to powers of two to be in line with most of the settings that already exist, as the ranks for LoRA are also chosen to be powers of $2$.
\subsection{Preliminaries}
In this section, we consider the default low-rank adaptation for a pretrained layer $W_0$ as $W = W_0 + s_R \cdot A B$, where $A \in \R^{m \times R}$ and $B \in \R^{R \times n}$ are the rank-$R$ adapters and the scaling factor $s_R$, with $s_z \in \{1, \frac{1}{z}, \frac{1}{\sqrt{z}}\}, \forall z \in \N \backslash \{0\}$. Given the input activation $x \in \R^{d \times m}$ for the current layer, the forward pass for LoRA is presented in Equation~\ref{eq:LoRA}. The fundamental difference between LoRA and DyLoRA is in the forward pass. Given an integer $k$ sampled from $S$ uniformly at random, the forward pass of DyLoRA is presented in Equation~\ref{eq:DyLoRA}.
\begin{align}
    \textbf{\textsc{LoRA: }} Y &= x(W_0 + s_R A B) \label{eq:LoRA} \\
    \textbf{\textsc{DyLoRA: }} Y &= x(W_0 + s_k A_k B_k) \label{eq:DyLoRA} 
\end{align}
\subsection{\method}
Our approach stores the same LoRA adapters $A \in \R^{m \times R}$ and $B \in \R^{R \times n}$ and instead of using the forward passes for LoRA and DyLoRa in Equations~\ref{eq:LoRA} and~\ref{eq:DyLoRA}, we include all slices $A_r, B_r$ into the forward pass as in Equation~\ref{eq:MatryoshkaLora}:
\begin{equation}
\textstyle
    \textbf{\method: }Y = x \left(W_0 + \sum_{r \in S} s_r A_r B_r\right) \label{eq:MatryoshkaLora} 
\end{equation}
The goal of \method is to train accurate hierarchical low-rank representations for different ranks inside the same LoRA adapters $A$ and $B$ such that the accuracy drop for ranks $r < R$ is minimized, as the loss will contain contributions from all ranks $r \in S$. Our goal is to update each slice $A_k$ and $B_k$ using gradient signal computed for the current batch of data. In contrast, DyLoRA uses the gradient from the current batch to update only the first $k$ columns/rows, while the rest $R-k$ do not receive any update, making it data-inefficient.

Equation~\ref{eq:MatryoshkaLora} cannot be implemented as it is in PyTorch because the framework does not allow propagating gradients only through a slice of parameter. Therefore, we must use the full adapters $A$ and $B$ and mask the gradients accordingly:
\begin{equation}
\textstyle
    Y = x\left(W_0 + \sum_{r \in S} s_r \cdot (A \odot M_r^A) (B \odot M_r^B)\right) \label{eq:FloatMasksMatryoshkaLoRA}
\end{equation}
where $M_r^A$ and $M_r^B$ are binary masks of the same shape as $A$ and $B$ where we set to $1$ only the first $r$ columns and first $r$ rows of $M^A$ and $M^B$, respectively. For example, for $m=n=R=3$ and $S = \{1,2\}$, we would have the following masks:
\begin{equation*}
    M_1^A = \begin{pmatrix}
        1 & 0 & 0 \\
        1 & 0 & 0 \\
        1 & 0 & 0 \\
    \end{pmatrix};
    M_2^A = \begin{pmatrix}
        1 & 1 & 0 \\
        1 & 1 & 0 \\
        1 & 1 & 0 \\
    \end{pmatrix};
    M_1^B = \begin{pmatrix}
        1 & 1 & 1 \\
        0 & 0 & 0 \\
        0 & 0 & 0 \\
    \end{pmatrix};
    M_2^B = \begin{pmatrix}
        1 & 1 & 1 \\
        1 & 1 & 1 \\
        0 & 0 & 0 \\
    \end{pmatrix}.
\end{equation*}


If we ignore the overhead of the element-wise multiplications between adapters $A$ and $B$ and their corresponding masks $M^A$ and $M^B$, then the Equation~\ref{eq:FloatMasksMatryoshkaLoRA} would require $|S|$ matmuls for each layer with inner dimensions $r \in S$, compared to one matmul for LoRA and DyLoRA with inner dimensions $R$ and $k$, which is clearly an undesired overhead.
During training, the boolean masks $M_r^A$ and $M_r^B$ must be stored and applied to the gradients computed for $A$ and $B$ at each forward pass. Even though they are boolean values, storing two masks per linear layer is against the simplicity of LoRA. Next, our goal is to simplify the formulation of \method. A careful inspection of Equation~\ref{eq:MatryoshkaLora} suggests there exist two matrices $C_A, C_B$ with the same shape as $A$ and $B$ such that:
\begin{equation}
\textstyle
    (A \odot C_A)(B \odot C_B) = \sum_{r \in S} (A \odot M_r^A) (B \odot M_r^B) \label{eq:SimplifiedMatryoshkaLora}
\end{equation}
In other words, we do not have to store individual boolean masks $M_r^{A,B}$ for $r \in S$ and potentially share these masks among layers whenever the dimensions allow. Instead, we can simply use two matrices $C_A,C_B$ per layer instead of $2 \cdot |S|$ matrices, which completely remove the need for masks $M_r^{A,B}$, as well as the need the loop in Equation~\ref{eq:MatryoshkaLora}. At this stage, we removed the boolean masks $M_r^{A,B}$ and we are left with the floating point matrices $C_A,C_B$, which still increase memory usage, even when stored in half precision.
We go one step further towards simplifying the forward pass for \method and observe that $C_A,C_B$ can be replaced with a diagonal matrix $P$. We provide more details in Appendix~\ref{appendix:simplification}. Therefore, we obtain the simplest form as:
\begin{equation}
    (A \odot C_A)(B \odot C_B) = A \cdot \text{diag}(P) \cdot B = (A * P) \cdot B, \label{eq:DiagMatryoshkaLora}
\end{equation}
where $P$ is an $R$-dimensional vector. Note we do not need to explicitly create the matrix $diag(P) \in \R^{R \times R}$. Instead, we can simply multiply the row $i$ of $A$ by $P_i$ to obtain the same effect at the lowest possible cost. The vector $P$ is the same for all layers and therefore the memory overhead of our \method is only the additional vector $P$ with $R$ elements. The final formulation of the forward pass for \method is based on the observation that there exists an $R$-dimensional vector $P_k$ such that $A_k \cdot B_k = A \cdot \text{diag}(P_k) \cdot B$, with $P_k = \text{diag}(\underbrace{1, \cdots, 1}_{k \text{ values}}, \underbrace{0, \cdots, 0}_{R-k \text{ values}})$ such that:
\begin{equation}
\textstyle
    Y = xW = x\left(W_0 + (A * P) B\right), \quad\text{with}\quad P = \sum_{r \in S} s_r \cdot P_r \label{eq:FinalMatryoshkaLora}
\end{equation}
Figure~\ref{fig:schematic} shows the schematic of \method: the diagonal matrix $P$ is used during the training using the forward pass in Equation~\ref{eq:FinalMatryoshkaLora}; during evaluation, we discard the matrix $P$, choose a specific rank $k$ and perform dynamic inference according to the default LoRA formula described in Equation~\ref{eq:LoRA}. In Appendix~\ref{appendix:theory} we provide a theoretical view of our formulation.

\subsubsection{How to compute $P$?}
\noindent
\begin{minipage}[t]{0.64\textwidth}
    In Algorithm~\ref{algorithm:vector-P} we show the steps to create the vector $P$ with $R$ components. For each rank $r$ from $R$ down to $1$, we use the integer $c$ to count how many times the sub-rank $k$ is employed in the sum in Equation~\ref{eq:MatryoshkaLora}. After that, the value for $P$ at component $r$ (denoted by $p_r$) is obtained as the sum of $s_{r_k}$, where $r_k$ is the rank at index $k$ in set $S$ containing training ranks, as shown in Equations~\ref{eq:P1},~\ref{eq:P2},~\ref{eq:P34} and~\ref{eq:P5678} as an example. Note that the summation loops through indices of $S$ indexed from $1$ to $|S|$ and not its elements. For example, $R=8$ and $S=\{1, 2, 4, 8\}$ will generate the vector $P$ with the following components $p_i$:
    \begin{alignat}{4}
    p_1       &= s_1 + &&s_2 + &&s_4 + &&s_8 \label{eq:P1}\\
    p_2       &= &&s_2 + &&s_4 + &&s_8 \label{eq:P2} \\
    p_3 = p_4     &= &&{} &&s_4 + &&s_8 \label{eq:P34} \\
    p_5 = p_6 = p_7 = p_8 &= &&{} &&{} &&s_8 \label{eq:P5678} 
    \end{alignat}
\end{minipage}
\hfill
\begin{minipage}[t]{0.34\textwidth}
    \vspace{-2em}
    \begin{algorithm}[H]
        \caption{Algorithm to compute vector $P$ in Equation~\ref{eq:DiagMatryoshkaLora} for our \method}\label{algorithm:vector-P}
        \begin{algorithmic}[1]
            \STATE \textbf{inputs:}
            \begin{itemize}[leftmargin=*]
                \item $S$ - set of training ranks
                \item $R$ - maximum rank of $A,B$
            \end{itemize}
            \STATE $P \gets 0_R$ \gray{// $R$ zeros}
            \STATE $c \gets 0$
            \FOR{$r = R$ down to $1$}
                \IF{$r \in S$}
                    \STATE $c \gets c+1$
                \ENDIF
                \STATE $p_r \gets \sum_{k=|S|-c}^{|S|} s_{r_k}$
            \ENDFOR
            \STATE \textbf{return} $P \in \R^R$
        \end{algorithmic}
    \end{algorithm}
\end{minipage}
Specifically, $p_r$ will have contributions from all ranks $r$ that appear as subscripts in its expression above. When we use $s_r=1, \forall r \in S$, we will obtain $P = [4, 3, 2, 2, 1, 1, 1, 1]$, which has the effect of scaling the first column of $A$ and first row of $B$ by 4 and so on, thus increasing the contribution of the gradient. Observe the last components of $P$ are all $1$, meaning that the columns/rows $5,6,7,8$ from $A$ and $B$ respectively will get gradient signal only from the rank-$8$ component. In short, the vector $P$ embeds the global contribution for each column/row in Equation~\ref{eq:MatryoshkaLora}. The vector $P$ induces the desired effect of learning hierarchical low-rank features in our \method.

\subsubsection{Adapter Scaling}
The default forward pass for LoRA uses $s_R = \nicefrac{1}{R}$ in Equation~\ref{eq:LoRA}, which accounts for the multiplication of $A$ and $B$ that have both the the inner dimension $R$. Prior work in the literature~\cite{rsLoRA} suggests the scaling should be $1/\sqrt{r}$ instead of $1/r$. However, in the context of \method, this scaling requires a much larger learning rate to learn the desired hieararchical low-rank features because larger ranks will have a smaller contribution. This is justified by the inequalities $1 \leq \sqrt{r} < r$ which implies $\nicefrac{1}{r} < \nicefrac{1}{\sqrt{r}} \leq 1$. Therefore, to minimize the size of our learning rate grid, we choose $s_k=1$ instead of $s_k \in \{1/r, 1/\sqrt{r}\}$ in the experiments for \method. We provide an empirical example of this phenomenon in Section~\ref{section:ablating-scaling-parameters}.

\subsubsection{General Framework for Recovering Existing LoRA Approaches}
In this section we show our Matryoshka framework is general enough to obtain both LoRA and DyLoRA by manipulating the expression of vector $P$. To recover LoRA described in Equation~\ref{eq:LoRA}, we have to choose $P = \nicefrac{1}{r}\cdot \boldsymbol{1}_R$, an $R$-dimensional vector that contains only the values $\nicefrac{1}{r}$. To recover DyLoRA described in Equation~\ref{eq:DyLoRA} for a sampled rank $k ~ S$, we have to choose $P = \nicefrac{1}{k} \cdot [\underbrace{1, \cdots, 1}_{k \text{ values}}, \underbrace{0, \cdots, 0}_{R-k \text{ values}}]$, where the first $k$ components are equal to $\nicefrac{1}{k}$ and the rest are zero.
\subsection{Gradient Computation for \method}
In this section we explain how inserting the constant matrix diag($P$) between $A$ and $B$ scales the gradients for $A$ and $B$, even though in the implementation we multiply the columns of $A$ with elements of vector $P$ for efficiency. Let $W \in \R^{m \times n}$ be the pretrained layer and $\Delta \in \R^{m \times n}$ be the upstream gradient for the current linear layer, for which we compute the gradients $\nabla_A$ and $\nabla_B$ for $A$ and $B$, respectively. Then, the gradients used in the optimizer for $A$ and $B$ have the following expression:
\vspace{-1em}
\begin{align}
    \nabla_A &= \Delta \cdot B^\top \cdot \text{diag}(P) \in \R^{m \times R}\label{eq:grad-A} \\
    \nabla_B &= \text{diag}(P) \cdot A^\top \cdot \Delta \in \R^{R \times n} \label{eq:grad-B}
\end{align}
Given the structure of diag($P$) described in previous sections, Equations~\ref{eq:grad-A} and~\ref{eq:grad-B} illustrate how the matrix diag($P$) scales the gradients $\nabla_A$ and $\nabla_B$, thus learning the hierarchical, low-rank representations we are targeting in \method.
\subsection{Evaluation Metrics}
Given a model fine-tuned with a LoRA variant (can be LoRA, DyLoRA or \method) and a set of evaluation ranks $S$, we define the set of accuracies $A_S = \{a_k \; | \; k \in S\}$ obtained when evaluating the model with rank $k \in S$ using the forward pass $Y = x (W_0 + s_k \cdot A_k B_k)$, regardless the underlying LoRA-type.

For example, if $A$ and $B$ have the bottleneck dimension $R = 16$ and we choose the evaluation ranks $k \in S = \{1, 2, 4, 8, 16\}$ on a specific dataset $\mathcal{D}$, we obtain a set of fine-tuning accuracies $A_S = \{a(1), a(2), a(4), a(8), a(16)\}$, where $a(k)$ is the accuracy on dataset $\mathcal{D}$ for rank $k$.

Given $S$ and $A_S$, we denote $r_i$ as the $i$-th rank in $S$ and $a(r_i)$ as the accuracy obtained with rank $r_i$, we compute two versions for an aggregate metric called the Area Under the Rank Accuracy Curve (AURAC), which employs the trapezoidal rule:
\begin{align}
    \text{AURAC} &= \frac{1}{r_{|S|} - r_1} \sum_{i=1}^{|S|-1} \frac{a(r_i) + a(r_{i+1})}{2} \cdot (r_{i+1} - r_i) \\
    \text{log-AURAC} &= \frac{1}{\log_2(r_{|S|}) - \log_2(r_1)} \sum_{i=1}^{|S|-1} \frac{a(r_i) + a(r_{i+1})}{2} \cdot \left(\log_2(r_{i+1}) - \log_2(r_i)\right)
\end{align}
The AURAC metric takes into account the distance between consecutive ranks in the set $S$ containing training ranks. For our example mentioned above, the accuracy area between ranks $8$ and $16$ will have the largest proportion in the resulting AURAC metric, equivalent to $(16 - 8) / (16 - 1) = 8 / 15 = 53.3\%$, which would bias the final AURAC value towards the accuracy of the largest interval between ranks $8$ and $16$, which is in line with the expectation that the LoRA adapters should learn hierarchical, low-rank features: a large value for the AURAC metric means all intermediary ranks achieve high accuracy. In our experiments we use AURAC by default.

The log-AURAC metric is designed only for the case when $S$ contains ranks that are power of two, which would result in all ranks being equally distant, therefore all intervals will contribute uniformly to the log-AURAC metric. For our example, all intervals will weighted by $1 / 4 = 25\%$. For the datasets we tested on, we haven't seen significant differences between AURAC and log-AURAT and therefore we chose the best runs based on the AURAC metric.

\vspace{-1em}
\section{Experimental Results} \label{section:experiments}
\textbf{Setting.} To prove the effectiveness of our \method in learning hierarchical low-rank features, we fine-tune Llama-3.2-1B-Instruct and Llama-3.1-8B-Instruct models on GSM-8k and OpenPlatypus~\cite{platypus} and test on GSM-8K ($3/8-$shot settings) and Open LLM Leaderboard~\cite{open-llm-leaderboard}, such as ARC-C~\cite{arc-challenge} and HellaSwag~\cite{hellaswag}, respectively.

\textbf{Ranks used.} We reiterate that the LoRA adapters in our work have shapes $A \in \R^{m \times R}$ and $B \in \R^{R \times n}$, where $R$ is the maximum rank (bottleneck size), which we ablate over in our experiments to assess the quality of each sub-rank. We choose the set $S$ that contains all powers of 2 up to the maximum rank $R$ (such as $32$ or $256$). Then, the ranks used for training and evaluation are determined by the set $S_R = \{1, 2, 4, ..., R\}$ (all powers of 2).

\textbf{Hyper-params \& Adapter Type.} We consider the best performing runs the ones achieving largest average AURAC metric, which we also report alongside evaluation accuracies for each particular rank. Averaging is performed across $3$ seeds ($42, 666, 2408$) and we omit the error bars or standard deviation for clarity reasons. We use $3$ epochs for all datasets in our study. To keep the compute costs for the evaluation low, we mention the grid we used to ablate over the learning rate in each section separately. All experiments in this section use Equation~\ref{eq:LoRA} for LoRA, Equation~\ref{eq:DyLoRA} for DyLoRA and Equation~\ref{eq:FinalMatryoshkaLora} for \method. We use the default AdamW~\cite{adamw} optimizer for fine-tuning.

\textbf{Hardware \& Software.} All our experiments are performed in single-GPU setting on Nvidia H100 GPUs with 80GB of RAM under PyTorch v$2.8.0$, \texttt{lm-eval} v$0.4.9.1$, \texttt{transformers} v$4.57.6$ and \texttt{datasets} v$3.6.0$.

\textbf{Overhead.} Since our approach adds one simple operation (multiplying the columns of adapter $A$ with vector $P$), the required resources in terms of running time and memory of \method are identical to LoRA and DyLoRA.

\subsection{Fine-Tuning Llama-3.2-1B-Instruct on GSM-8K for different ranks $R$}\label{section:llama-1B-gsm8k-ranks-1-to-32}
The first set of experiments focuses on fine-tuning Llama-3.2-1B-Instruct~\cite{llama3} on GSM-8k~\cite{gsm8k} using LoRA, DyLoRA and \method. We fix the bottleneck size to $R=32$ and use the ranks in the set $S_{32}=\{1, 2, 4, 8, 16, 32\}$ to train the models and evaluate them using 8-shot strategy via lm-evaluation-harness~\cite{lm-eval-harness}.

We perform grid search over learning rates $\eta \in \{1e-5, 2e-5, 4e-5, 6e-5, 8e-5\}$ and report the results in Table~\ref{section:llama-1B-gsm8k-ranks-1-to-32}. For the discussion that follows, we mention that the baseline accuracy of the pre-trained model (before any fine-tuning) is $34.7\%$.

\textbf{Rank-wise Accuracy.} Our \method dominates across the entire rank spectrum. While LoRA and DyLoRA exhibit relatively modest gains as rank increases, typically pleateauing in the mid $34\%$ range, our approach scales more effectively with rank and exhibits a significant increase in evaluation accuracy starting with $R=2$. At higher ranks, such as $r \in \{16, 32\}$, \method achieves accuracies up to $39\%$ and $38\%$, respectively, clearly surpassing the best LoRA ($\approx 35\%$) and DyLoRA ($\approx 35.5\%$).

We would like to emphasize that the evaluation accuracy obtained for the sub-rank $r=1$ obtained when trained with bottleneck size $R=32$ is better already better than any other sub-rank for LoRA and DyLoRA. When we compare against the evaluation accuracy of $34.7\%$ for the pre-trained model, we observe that fine-tuning with LoRA and DyLoRA does not improve accuracy significantly. Moreover, only deploying $k=32$ obtained for $R=32$ would lead to better performance compared to the pre-trained model.

\textbf{AURAC.} In terms of AURAC, which captures the performance aggregated across ranks, our method provides again a clear advantage. Both LoRA and DyLoRA achieve peak AURAC values of $\approx 35\%$, whereas \method reaches $38.4\%$, representing a substantial improvement. This gap highlights both stronger peak performance and more consistent gains across all ranks.

\textbf{Key takeaway.} Selecting rank $r\in\{4, 8\}$ with accuracies $37-39\%$) from \method trained with bottleneck size $R=32$ (the very last row of Table~\ref{table:llama-1B-gsm8k-ranks-1-to-32}) yields higher accuracy compared to any sub-rank chosen from either LoRA or DyLoRA.

\begin{table}[h]
\caption{[Results for Section~\ref{section:llama-1B-gsm8k-ranks-1-to-32}] AURAC metric and per-rank, evaluation accuracies for Llama-3.2-1B-Instruct on GSM-8k (8-shot) across multiple ranks. \textbf{We mark in bold the best AURAC per row} and emphasize the accuracy of pre-trained model (before finetuning) of $34.7\%$.}
\label{table:llama-1B-gsm8k-ranks-1-to-32}
\centering
\small
\begin{sc} 
\setlength{\tabcolsep}{3pt}
\begin{tabular}{lccccccccr}
\toprule
\multirow{2}{*}{\makecell{Adapter\\Type}} & \multicolumn{6}{c}{Eval Accuracy per Rank} & \multirow{2}{*}{AURAC}\\
\cmidrule(lr){2-7}
           & 1    & 2    & 4    & 8    & 16   & 32   &      & \\
\midrule
\multirow{6}{*}{LoRA}   &  32.4 &      &      &      &      &      & \textbf{32.4} \\
                        & 32.4 & 33.6 &      &      &      &      & 33.0 \\
                        & 33.5 & 34.8 & 34.3 &      &      &      & 34.4 \\
                        & 34.2 & 34.1 & 34.6 & 34.9 &      &      & 34.5 \\
                        & 32.5 & 33.6 & 34.5 & 34.8 & 34.9 &      & 34.6 \\
                        & 33.3 & 34.3 & 34.3 & 34.3 & 34.5 & 34.6 & 34.5 \\
\midrule
\multirow{6}{*}{DyLoRA} & 32.4 &      &      &      &      &      & \textbf{32.4} \\
                        & 32.9 & 33.5 &      &      &      &      & 33.2 \\
                        & 33.4 & 34.0 & 34.3 &      &      &      & 34.0 \\
                        & 33.1 & 33.8 & 34.6 & 34.7 &      &      & 34.4 \\
                        & 32.9 & 33.9 & 34.6 & 34.8 & 34.8 &      & 34.6 \\
                        & 34.4 & 34.5 & 34.1 & 34.7 & 34.9 & 35.4 & 34.9 \\
\midrule
\multirow{6}{*}{\method} & 32.3 &      &      &      &      &      & 32.3 \\
                      & 35.9 & 34.7 &      &      &      &      & \textbf{35.3} \\
                      & 36.5 & 36.8 & 35.4 &      &      &      & \textbf{36.3} \\
                      & 35.2 & 36.0 & 36.4 & 37.3 &      &      & \textbf{36.5} \\
                      & 35.3 & 35.9 & 37.4 & 37.7 & 37.1 &      & \textbf{37.2} \\
                      & 35.8 & 35.6 & 37.2 & 38.8 & 39.1 & 38.3 & \textbf{38.4} \\
\bottomrule
\end{tabular}
\end{sc}
\end{table}

\subsection{Fine-Tuning Llama-3.1-8B-Instruct on GSM-8K}\label{section:llama-8B-gsm8k-3-8-shots}
The second set of experiments focuses on fine-tuning Llama-3.1-8B-Instruct~\cite{llama3} on GSM-8k~\cite{gsm8k} using the same LoRA, DyLoRA and \method and evaluated using 3/8-shots strategies. We fix the bottleneck size to $R=256$ and use the ranks in the set $S_{256} = \{1, 2, 4, 8, 16, 32, 64, 128, 256\}$ for both training and evaluation via the same lm-evaluation-harness~\cite{lm-eval-harness}.

We perform grid search over learning rates $\eta \in \{5e-6, 6e-6, 7e-6, 8e-6, 9e-6, 1e-5, 2e-5, 3e-5, 4e-5, 5e-5\}$ and report the results in the upper half of Table~\ref{table:llama-8B-gsm8k-3-8-shots-platypus-25-10-shots}. We mention the accuracy of the pre-trained model (before any fine-tuning) in the \textbf{PT} column of the table.

\textbf{3-shot setting.} Our method consistently outperforms both LoRA and DyLoRA across nearly all ranks. While the evaluation accuracy for these two baselines fluctuate in the range $74-75\%$, \method achieves higher and more stable per-rank performance, reaching up to $78.6\%$ for rank $r=128$, an improvement of over $4\%$ compared to both pre-trained model and baselines. Moreover, our improvements are visible from low to high ranks, which is the purpose of the hierarchical low-rank representations. The improvements in per-rank accuracies translate to the AURAC score, where \method reaches more than $3\%$ compared to baselines.
\textbf{8-shot setting.} The advantage of \method remains consistent, reaching $80\%$ accuracy for ranks $r \in \{32, 64, 128\}$. However the gap decreases when $r$ increases. Our explanation is that more in-context information helps learning overall and shadows the influence of the adapter.

\textbf{Drops from 128 to 256.} In both 3/8-shot settings, there is a drop in accuracy when going from rank $128$ to $256$ for both LoRA and \method. We believe this happens because the bottleneck size $R=256$ might require more than $3$ epochs to fix the drop.

\textbf{Key takeaway.} \method closes the gap between 3-shot and 8-shot: if the goal is to achieve $77\%$ accuracy with 3-shots, one should pick the adapters trained with \method for $R=256$ and use $r=32$ or $r=64$ with 3-shots. In principle, would translate to shorter contexts compared to 8-shots and therefore lower inference costs.
\begin{table}[h]
\caption{[Results for Section~\ref{section:llama-8B-gsm8k-3-8-shots} and~\ref{section:llama-8B-platypus}] AURAC metric and per-rank evaluation accuracies for GSM-8k (3/8-shot) and Open LLM Leaderboard on ARC-C (25-shot) and Hellaswag (10-shot) for Llama-3.1-8B-Instruct on GSM-8k across multiple ranks. \textsc{PT} stands for the accuracy of the \textbf{P}re-\textbf{T}rained model (before fine-tuning) with $N$-shots.}
\label{table:llama-8B-gsm8k-3-8-shots-platypus-25-10-shots}
\centering
\small
\begin{sc} 
\setlength{\tabcolsep}{3pt}
\begin{tabular}{lccccccccccccr}
\toprule
\multirow{2}{*}{N} & \multirow{2}{*}{\makecell{Adapter\\Type}} & \multirow{2}{*}{PT} & \multicolumn{9}{c}{Eval Accuracy per Rank} & \multirow{2}{*}{AURAC}\\
\cmidrule(lr){4-12}
& & & 1 & 2 & 4 & 8 & 16 & 32 & 64 & 128 & 256 & & \\ 
\midrule
  & LoRA       &      & 74.9 & 74.8 & 74.1 & 74.2 & 75.0 & 73.6 & 74.7 & 74.6 & 73.6 & 74.3 \\ 
\multirow{2}{*}{\makecell{GSM-8k\\3-shot}} & DyLoRA     & 74.0 & 74.7 & 75.0 & 74.7 & 74.5 & 73.5 & 74.1 & 74.5 & 74.5 & 74.5 & 74.4 \\ 
  & Matryoshka &      & 73.5 & 73.5 & 74.7 & 75.0 & 75.9 & 76.8 & 77.6 & 78.6 & 76.4 & \textbf{77.4} \\
  
\midrule

  & LoRA       &      & 78.6 & 78.4 & 79.1 & 78.6 & 78.5 & 78.7 & 78.9 & 79.5 & 79.1 & \textbf{79.1} \\ 
\multirow{2}{*}{\makecell{GSM-8k\\8-shot}} & DyLoRA     & 77.8 & 78.5 & 78.5 & 77.9 & 78.6 & 78.5 & 78.4 & 78.8 & 79.5 & 79.8 & \textbf{79.2} \\ 
  & Matryoshka &      & 77.9 & 77.7 & 79.2 & 79.1 & 79.8 & 80.2 & 80.5 & 80.2 & 78.1 & \textbf{79.7} \\

\midrule

& LoRA &      & 56.7 & 56.9 & 56.6 & 56.9 & 56.9 & 56.9 & 57.1 & 57.1 & 57.0 & 57.0 \\
\multirow{2}{*}{\makecell{ARC-C\\25-shot}} & DyLoRA & 56.7 & 56.9 & 56.7 & 57.0 & 57.0 & 56.9 & 57.0 & 57.0 & 57.0 & 57.0 & 57.0 \\
& Matryoshka & & 56.9 & 56.8 & 56.9 & 57.0 & 57.2 & 57.3 & 57.6 & 58.4 & 58.1 & \textbf{58.0} \\

\midrule

& LoRA & & 59.0 & 59.1 & 59.2 & 59.1 & 59.2 & 59.2 & 59.2 & 59.2 & 59.1 & 59.2 \\
\multirow{2}{*}{\makecell{HellaSwag\\10-shot}} & DyLoRA & 59.3 & 59.2 & 59.2 & 59.2 & 59.2 & 59.1 & 59.2 & 59.2 & 59.2 & 59.2 & 59.2 \\
& Matryoshka & & 59.4 & 59.3 & 59.3 & 59.4 & 59.6 & 59.9 & 60.5 & 61.4 & 62.8 & \textbf{61.4} \\

\bottomrule
\end{tabular}
\end{sc}
\end{table}




\subsection{Fine-Tuning Llama-3.1-8B-Instruct on Open Platypus} \label{section:llama-8B-platypus}
The third set of experiments focuses on fine-tuning Llama-3.1-8B-Instruct~\cite{llama3} on Open Platypus~\cite{platypus} and evaluating on two datasets from Open LLM Leaderboard~\cite{open-llm-leaderboard}: ARC-Challenge~\cite{arc-challenge} ($25$-shots) and Hellaswag~\cite{hellaswag} ($10$-shots). The setting for $R$ is identical to Section~\ref{section:llama-8B-gsm8k-3-8-shots}. 

We perform grid search over learning rates $\eta \in \{6e-6, 7e-6, 8e-6, 9e-6, 1e-6, 2e-5, 3e-5, 4e-5\}$ and report the results in the lower half of Table~\ref{table:llama-8B-gsm8k-3-8-shots-platypus-25-10-shots}.

\textbf{ARC-C.} LoRA and DyLoRA exhibit an identical behavior, with both per-rank accuracies and AURAC clustered around $57\%$, a small improvement over the pre-trained accuracy of $56.7\%$. In contrast, \method achieves more than $58\%$ per-rank accuracy and $58\%$ AURAC, showing consistent imprvements as the rank increases.

\textbf{HellaSwag.} The advantage of our method is more pronounced for this dataset. While LoRA and DyLoRA remain effectively flat around the accuracy of the pre-trained model of $59.2\%$, \method exhibits steady gains with increasing rank, achieving $60.5\%$ for $r=64$, $61.4\%$ for $r=128$ and peaking at $62.8\%$ for $r=256$, which represents an increase of more than $3\%$ over the rank $r=256$ for LoRA and DyLoRA. In line with the increased rank-wise accuracy, AURAC score reaches $61.4\%$.

\subsection{Ablation for scaling parameters $s_k \in \{1, \nicefrac{1}{k}, \nicefrac{1}{\sqrt{k}}\}$} \label{section:ablating-scaling-parameters}
Our last set of experiments focuses on understanding the effect of the scale $s_k \in \{1, \nicefrac{1}{k}, \nicefrac{1}{\sqrt{k}}\}$ over the learning rate and evaluation accuracy/AURAC for our \method for a fixed bottleneck size $R=32$. Our hypothesis is that $s_k = \nicefrac{1}{\sqrt{r}}$ requires larger learning rate to achieve highest accuracy over a learning rate grid $G_\eta$.

We fine-tune Llama-3.2-1B-Instruct on GSM-8k with $R=32$ using three seeds and ablating over a learning rate grid $G_\eta \in \{5e-5, \cdots, 1e-3\}$, containing $14$ values. We would like to emphasize that changing $R$ directly influences the behavior of this experiment and because of that we decide to keep $R$ fixed and ablate only over the learning rate $\eta$.

We show the results of this experiment in Table~\ref{table:llama-1B-gsm8k-scale-ablation}, where we can see that our hypothesis is confirmed: compared to scaling $s_k=1$, we need a learning rate that is $9\times$ larger when we use scaling $s_k = \nicefrac{1}{\sqrt{k}}$ and $4\times$ larger when we use scaling $s_k = \nicefrac{1}{k}$.
\begin{table}[h]
\caption{[Results for Section~\ref{section:ablating-scaling-parameters}] Scaling ablation for \method on Llama-3.2-1B-Instruct on GSM-8k(8-shot). The accuracy of the pre-trained model (before fine-tuning) is $34.7\%$.}
\label{table:llama-1B-gsm8k-scale-ablation}
\centering
\small
\begin{sc} 
\setlength{\tabcolsep}{3pt}
\begin{tabular}{lcccccccccr}
\toprule
\multirow{2}{*}{\makecell{Scaling\\Type}} & \multicolumn{6}{c}{Eval Accuracy per Rank} & \multirow{2}{*}{AURAC} & \multirow{2}{*}{$\eta$}\\
\cmidrule(lr){2-7}
                               & 1    & 2    & 4    & 8    & 16   & 32   &   &\\
\midrule
       $s_k = \nicefrac{1}{r}$ & 36.0 & 37.2 & 36.8 & 36.5 & 38.3 & 38.8 & 37.8 & $9 \cdot 10^{-4}$ \\
\midrule
$s_k = \nicefrac{1}{\sqrt{r}}$ & 36.4 & 36.9 & 38.2 & 38.4 & 39.1 & 39.0 & 38.7 & $3 \cdot 10^{-4}$ \\
\midrule
                     $s_k = 1$ & 35.5 & 36.2 & 37.7 & 38.7 & 39.2 & 37.8 & 38.4 & $1 \cdot 10^{-4}$ \\
\bottomrule
\end{tabular}
\end{sc}
\end{table}
\section{Conclusion, Limitations and Broader Impact}\label{section:conclusion-limitations-impact}
\textbf{Conclusion.} We propose \method, a framework that modifies standard LoRA by inserting a carefully crafted diagonal matrix diag($P$) between adapters $A$ and $B$ that enables learning hierarchical, low-rank features inside the same adapters $A$ and $B$ as embedded prefixes. We demonstrate the effectiveness of this simple approach on Llama models with 1 billion and 8 billions of parameters on a few popular datasets in the literature. We also show that we can obtain LoRA and DyLoRA in our framework by simply changing the expression of the vector $P$. In addition propose a metric to assess the performance of a LoRA-like model when evaluated on multiple ranks to facilitate choosing a best model during the hyper-parameter tuning process.

\textbf{Limitations.} Despite this method being simple and having the same overhead as LoRA, the preparation of this work requires many model evaluations for each rank, which increases the actual runtime to assess the performance. Therefore, we believe our results can be improved by further hyper-parameter tuning. Specifically, one might experiment with, just to name a few, a wider grid for the learning rate, different number of epochs or weight decay. One more aspect worth mentioning is that our evaluations are performed with the same rank $k$ for the entire network. However, in our paper we haven't experimented with settings where different layers get different ranks for the same forward pass, as it is the case for studies in the literature that measure sensitivity of each layer which affects its LoRA rank.

\textbf{Broader Impact.} Our \method introduces a general framework in the LoRA-like literature that allows learning accurate, hierarchical low-rank features that live inside the same adapters, allowing more efficient deployment. However, it is important to note that while the purpose of our our technique is to reduce overall deployment costs, we do not have control over the applications where our method might be used in.

\section*{Acknowledgments}
We would like to thank the Scientific Computing Department at ISTA for providing access to computational resources to develop this work. MS’s work was supported by Research England under the Expanding Excellence in England (E3) funding stream, which was awarded to MARS:
Mathematics for AI in Real-world Systems in the School of Mathematical Sciences at Lancaster University.

\medskip

\printbibliography

\clearpage
\appendix

\section*{Appendix}
\tableofcontents
\clearpage

\section{Simplification of \method}\label{appendix:simplification}
In this section we explain how we simplify the expression of \method presented in Equation~\ref{eq:MatryoshkaLora}, where we choose scaling $s_r = 1$ for simplicity. Let's take an example of two LoRA adapters, where $A \in (3, 3)$ and $B \in (3, 3)$, from which we want to use ranks $r \in \{1,2,3\}$ and denote by $A_r, B_r$ the first $r$ columns for $A$ and first $r$ rows for $B$.
 
\[
    A = A_3 =
    \begin{pmatrix}
        a_{11} & a_{12} & a_{13} \\
        a_{21} & a_{22} & a_{23} \\
        a_{31} & a_{32} & a_{33}
    \end{pmatrix}
    \quad
    A_1 =
    \begin{pmatrix}
        a_{11} \\
        a_{21} \\
        a_{31}
    \end{pmatrix}
    \quad
    A_2 =
    \begin{pmatrix}
        a_{11} & a_{12} \\
        a_{21} & a_{22} \\
        a_{31} & a_{32}
    \end{pmatrix}
\]
\[
    B = B_3 =
    \begin{pmatrix}
        b_{11} & b_{12} & b_{13} \\
        b_{21} & b_{22} & b_{23} \\
        b_{31} & b_{32} & b_{33}
    \end{pmatrix}
    \quad
    B_1 =
    \begin{pmatrix}
        b_{11} & b_{12} & b_{13}
    \end{pmatrix}
    \quad
    B_2 =
    \begin{pmatrix}
        b_{11} & b_{12} & b_{13} \\
        b_{21} & b_{22} & b_{23}
    \end{pmatrix}
\]

We want to find matrices $C_A, C_B$ of the same shape as $A$ and $B$ such that $(A \odot C_A) (B \odot C_B) = \sum_r A_r B_r$. First, let's look at the result of multiplications $A_r \cdot B_r$ for each $r \in \{1, 2\}$ by explicitly writing the scalar products between rows of $A_r$ and columns of $B_r$

\[
    A_1 \cdot B_1 =
    \begin{pmatrix}
        a_{11}b_{11} & a_{11}b_{12} & a_{11}b_{13} \\
        a_{21}b_{11} & a_{21}b_{12} & a_{21}b_{13} \\
        a_{31}b_{11} & a_{31}b_{12} & a_{31}b_{13}
    \end{pmatrix}
    \quad
    A_2 \cdot B_2 =
    \begin{pmatrix}
    a_{11}b_{11} + a_{12}b_{21} & a_{11}b_{12} + a_{12}b_{22} & a_{11}b_{13} + a_{12}b_{23} \\
    a_{21}b_{11} + a_{22}b_{21} & a_{21}b_{12} + a_{22}b_{22} & a_{21}b_{13} + a_{22}b_{23} \\
    a_{31}b_{11} + a_{32}b_{21} & a_{31}b_{12} + a_{32}b_{22} & a_{31}b_{13} + a_{32}b_{23}
    \end{pmatrix}
\]
\[
    A_3 \cdot B_3 =
    \begin{pmatrix}
    a_{11}b_{11} + a_{12}b_{21} + a_{13}b_{31} & a_{11}b_{12} + a_{12}b_{22} + a_{13}b_{32} & a_{11}b_{13} + a_{12}b_{23} + a_{13}b_{33} \\ 
    a_{21}b_{11} + a_{22}b_{21} + a_{23}b_{31} & a_{21}b_{12} + a_{22}b_{22} + a_{23}b_{32} & a_{21}b_{13} + a_{22}b_{23} + a_{23}b_{33} \\
    a_{31}b_{11} + a_{32}b_{21} + a_{33}b_{31} & a_{31}b_{12} + a_{32}b_{22} + a_{33}b_{32} & a_{31}b_{13} + a_{32}b_{23} + a_{33}b_{33}
    \end{pmatrix}
\]
\[
    \sum_{r \in \{1,2,3\}} A_r B_r =
    \begin{pmatrix}
    \bs{3}a_{11}b_{11} + \bs{2}a_{12}b_{21} + \bs{1}a_{13}b_{31} & \bs{3}a_{11}b_{12} + \bs{2}a_{12}b_{22} + \bs{1}a_{13}b_{32} & \bs{3}a_{11}b_{13} + \bs{2}a_{12}b_{23} + \bs{1}a_{13}b_{33} \\
    \bs{3}a_{21}b_{11} + \bs{2}a_{22}b_{21} + \bs{1}a_{23}b_{31} & \bs{3}a_{21}b_{12} + \bs{2}a_{22}b_{22} + \bs{1}a_{23}b_{32} & \bs{3}a_{21}b_{13} + \bs{2}a_{22}b_{23} + \bs{1}a_{23}b_{33} \\
    \bs{3}a_{31}b_{11} + \bs{2}a_{32}b_{21} + \bs{1}a_{33}b_{31} & \bs{3}a_{31}b_{12} + \bs{2}a_{32}b_{22} + \bs{1}a_{33}b_{32} & \bs{3}a_{31}b_{13} + \bs{2}a_{32}b_{23} + \bs{1}a_{33}b_{33}
    \end{pmatrix}\\
\]

By visual inspection, we observe the elements from the first row of $A$ and first column of $B$ are scaled by $3$, which is the total number of ranks we test for. The second row/column is scaled by $2$ and finally, the third row/column is scaled by $1$. Since we need to make sure we learn hierarchical features for both $A$ and $B$, we can use the square root of coefficients $3,2,1$ for $C_A$ and $C_B$:

\[
    C_A = 
    \begin{pmatrix}
        \sqrt{3} & \sqrt{2} & 1 \\
        \sqrt{3} & \sqrt{2} & 1 \\
        \sqrt{3} & \sqrt{2} & 1
    \end{pmatrix}
    \quad
    C_B = 
    \begin{pmatrix}
        \sqrt{3} & \sqrt{3} & \sqrt{3} \\
        \sqrt{2} & \sqrt{2} & \sqrt{2} \\
        1 & 1 & 1
    \end{pmatrix}
\]

Moreover, instead of using matrices $C_A$ and $C_B$, we can obtain the matrix $\sum_{r \in \{1,2,3\}} A_r B_r$ by multiplying column $i$ of $A$ by $P_i$, thus yielding the final \method forward pass during training as in Equation~\ref{eq:FinalMatryoshkaLora}. Specifically, the product $A * P$ yields the following matrix:
\[
    A * P =
    \begin{pmatrix}
        \bs{3} a_{11} & \bs{2} a_{12} & \bs{1} a_{13} \\
        \bs{3} a_{21} & \bs{2} a_{22} & \bs{1} a_{23} \\
        \bs{3} a_{31} & \bs{2} a_{32} & \bs{1} a_{33}
    \end{pmatrix}
\]


\section{DyLoRA vs MatryoshkaLoRA: Theory View}\label{appendix:theory}

Consider a linear layer with pretrained weights $W_0\in\mathbb{R}^{m\times n}$. Standard LoRA introduces trainable adapters $A\in\mathbb{R}^{m\times R}$ and $B\in\mathbb{R}^{R\times n}$, and replaces the layer by
$$
W = W_0 + s AB,
$$
where $s=s_R$ is a rank-dependent scaling factor, for example $s_R=\nicefrac{\alpha}{R}$ or $s_R=\nicefrac{\alpha}{\sqrt{R}}$. Given a fine-tuning dataset and loss function $\ell$, the fine-tuning objective is
$$
\min_{A,B} \ell(W_0 + sAB).
$$

The limitation of the standard approach is that fine-tuning is tied to a specific rank $R$. If, after training, we want to use a smaller rank $r<R$, then in general we would need to fine-tune again. A naive alternative is to take the first $r$ columns of $A$ and the first $r$ rows of $B$, but there is no reason for these truncated adapters to perform well unless the training procedure explicitly encourages this structure.

Let $R$ be the maximum rank, and let
$$
{\cal S}=\{1\le r_1<r_2<\dots<r_k\le R\}
$$
be the collection of smaller ranks we would like to support. Our goal is to fine-tune a single pair of rank-$R$ adapters $A,B$ such that, for every $r\in{\cal S}$, the truncated adapters
$$
A_r = A[:,1\!:\!r], \qquad B_r = B[1\!:\!r,:]
$$
define a useful rank-$r$ adaptation $W_0 + s_r A_rB_r$. This leads naturally to the family of objectives
$$
L_r(A,B) \;=\; \ell(W_0 + s_r A_r B_r), \qquad r\in{\cal S},
$$
which we would like to optimize simultaneously.
A natural way to train one adapter pair for all ranks is the weighted multi-rank objective
\[
\min_{A,B}L_{\rm multi}(A,B),
\]
where
\[
L_{\rm multi}(A,B)
=
\sum_{r\in{\cal S}}\lambda_rL_r(A,B)
=
\sum_{r\in{\cal S}}\lambda_r\,
\ell(W_0+s_rA_rB_r),
\]
with normalized weights
\[
\lambda_r\ge 0,
\qquad
\sum_{r\in{\cal S}}\lambda_r=1.
\]
The weights \(\lambda_r\) specify how much training emphasis is placed on each supported rank. Now define the diagonal truncation matrix
\[
P_r
=
{\rm diag}(
\underbrace{1,\dots,1}_{r},
\underbrace{0,\dots,0}_{R-r}
)
\in\mathbb{R}^{R\times R}.
\]
Then $A_rB_r=AP_rB$, and therefore
\[
L_r(A,B)
=
\ell(W_0+s_rAP_rB).
\]
Hence the multi-rank objective can also be written as
\[
L_{\rm multi}(A,B)
=
\sum_{r\in{\cal S}}\lambda_r\,
\ell(W_0+s_rAP_rB).
\]

\subsection{DyLoRA as stochastic optimization of the multi-rank objective}

DyLoRA studies the stochastic version of the multi-rank objective above. In DyLoRA, one chooses a rank range
\[
{\cal S}
=
\{r_{\min},r_{\min}+1,\dots,r_{\max}\},
\]
and samples a rank $b\sim p_B(\cdot)$ at each training step. The LoRA matrices are then truncated to
\[
A_b=A[:,1\!:\!b],
\qquad
B_b=B[1\!:\!b,:],
\]
and the forward pass uses the rank-\(b\) perturbation $\Delta_b = s_bA_bB_b$ in the usual DyLoRA scaling $s_b=\frac{\alpha}{b}$. Thus the sampled DyLoRA loss at one training step is
$$
L_b(A,B)
=
\ell(W_0+s_bA_bB_b).
$$

Taking expectation over the sampled rank gives
\[
\mathbb{E}_{b\sim p_B}\big[L_b(A,B)\big]
=
\sum_{r\in{\cal S}}p_B(r)\,
\ell(W_0+s_rA_rB_r).
\]
Therefore DyLoRA corresponds to the multi-rank objective with $\lambda_r=p_B(r)$. That is,
\[
L_{\rm multi}(A,B)
=
\mathbb{E}_{b\sim p_B}
\left[
L_b(A,B)
\right].
\]

Equivalently, using the truncation matrices \(P_r\),
\[
L_b(A,B)
=
\ell(W_0+s_bAP_bB),
\]
and
\[
L_{\rm multi}(A,B)
=
\mathbb{E}_{b\sim p_B}
\left[
\ell(W_0+s_bAP_bB)
\right]
=
\sum_{r\in{\cal S}}\lambda_r
\ell(W_0+s_rAP_rB).
\]

For DyLoRA update, the sampled gradient is an unbiased estimator of the full multi-rank gradient:
\[
\mathbb{E}_{b\sim p_B}
\left[
\nabla_{A,B}L_b(A,B)
\right]
=
\nabla_{A,B}L_{\rm multi}(A,B).
\]
So DyLoRA does not need to compute the full sum over all ranks at each step. Instead, it solves the same expected objective by sampling one rank-loss term at a time, exactly like SGD solves a full-data objective by sampling minibatches.

\subsection{From DyLoRA's stochastic objective to a MatryoshkaLoRA surrogate}

We now connect the stochastic DyLoRA objective to a deterministic MatryoshkaLoRA-style forward pass. Define the loss as a function of the LoRA perturbation:
\[
f(\Delta)=\ell(W_0+\Delta).
\]
For each \(r\in{\cal S}\), define $\Delta_r=s_rA_rB_r=s_rAP_rB$. Then the multi-rank objective is
\[
L_{\rm multi}(A,B)
=
\sum_{r\in{\cal S}}\lambda_rf(\Delta_r).
\]
Assume now that $f$ is differentiable and $L$-smooth, i.e., there exists a constant $L>0$ such that
\[
\|\nabla f(\Delta)-\nabla f(\Delta')\|_{\rm F}
\le
L\|\Delta-\Delta'\|_{\rm F}
\]
for all perturbations \(\Delta,\Delta'\). This implies the first-order expansion
\[
f(\Delta)
=
f(0)+\langle \nabla f(0),\Delta\rangle+{\cal R}(\Delta),
\]
where $|{\cal R}(\Delta)|\le\frac{L}{2}\|\Delta\|_{\rm F}^2$. Therefore,
\begin{align*}
\sum_{r\in{\cal S}}\lambda_rf(\Delta_r)
&=
\sum_{r\in{\cal S}}\lambda_r
\Big(
f(0)
+
\langle \nabla f(0),\Delta_r\rangle
+
{\cal R}(\Delta_r)
\Big) \\
&=
f(0)
+
\left\langle
\nabla f(0),
\sum_{r\in{\cal S}}\lambda_r\Delta_r
\right\rangle
+
\sum_{r\in{\cal S}}\lambda_r{\cal R}(\Delta_r).
\end{align*}
On the other hand,
\[
f\left(\sum_{r\in{\cal S}}\lambda_r\Delta_r\right)
=
f(0)
+
\left\langle
\nabla f(0),
\sum_{r\in{\cal S}}\lambda_r\Delta_r
\right\rangle
+
{\cal R}
\left(
\sum_{r\in{\cal S}}\lambda_r\Delta_r
\right).
\]
Subtracting the two identities gives
\[
\sum_{r\in{\cal S}}\lambda_rf(\Delta_r)
-
f\left(\sum_{r\in{\cal S}}\lambda_r\Delta_r\right)
=
\sum_{r\in{\cal S}}\lambda_r{\cal R}(\Delta_r)
-
{\cal R}
\left(
\sum_{r\in{\cal S}}\lambda_r\Delta_r
\right).
\]
Hence,
\begin{align*}
\left|
\sum_{r\in{\cal S}}\lambda_rf(\Delta_r)
-
f\left(\sum_{r\in{\cal S}}\lambda_r\Delta_r\right)
\right|
&\le
\sum_{r\in{\cal S}}\lambda_r|{\cal R}(\Delta_r)|
+
\left|
{\cal R}
\left(
\sum_{r\in{\cal S}}\lambda_r\Delta_r
\right)
\right| \\
&\le
\frac{L}{2}
\sum_{r\in{\cal S}}\lambda_r\|\Delta_r\|_{\rm F}^2
+
\frac{L}{2}
\left\|
\sum_{r\in{\cal S}}\lambda_r\Delta_r
\right\|_{\rm F}^2.
\end{align*}
By convexity of the squared norm,
\[
\left\|
\sum_{r\in{\cal S}}\lambda_r\Delta_r
\right\|_{\rm F}^2
\le
\sum_{r\in{\cal S}}\lambda_r\|\Delta_r\|_{\rm F}^2.
\]
Therefore,
\[
\left|
\sum_{r\in{\cal S}}\lambda_rf(\Delta_r)
-
f\left(\sum_{r\in{\cal S}}\lambda_r\Delta_r\right)
\right|
\le
L
\sum_{r\in{\cal S}}\lambda_r\|\Delta_r\|_{\rm F}^2.
\]
Equivalently,
\[
\sum_{r\in{\cal S}}\lambda_rf(\Delta_r)
=
f\left(\sum_{r\in{\cal S}}\lambda_r\Delta_r\right)
+
O\left(
\sum_{r\in{\cal S}}\lambda_r\|\Delta_r\|_{\rm F}^2
\right).
\]

Using \(A_rB_r=AP_rB\), we obtain
\begin{align*}
\ell
\left(
W_0+
\sum_{r\in{\cal S}}\lambda_rs_rA_rB_r
\right)
&=
\ell
\left(
W_0+
\sum_{r\in{\cal S}}\lambda_rs_rAP_rB
\right) \\
&=
\ell
\left(
W_0+
A
\left[
\sum_{r\in{\cal S}}\lambda_rs_rP_r
\right]
B
\right).
\end{align*}
Then the deterministic surrogate objective is
\[
\ell(W_0+APB), 
\quad P
=
\sum_{r\in{\cal S}}\lambda_rs_rP_r.
\]

This is the MatryoshkaLoRA-style surrogate: instead of sampling one rank per step as in DyLoRA, we use a single forward pass with a weighted combination of all nested rank components.

\subsection{Summary of the connection}

The connection can be summarized as follows. DyLoRA defines a sampled-rank training procedure:
\[
b\sim p_B(\cdot),
\qquad
L_b(A,B)
=
\ell(W_0+s_bA_bB_b).
\]
With $\lambda_r=p_B(r)$, this is stochastic optimization of
\[
L_{\rm multi}(A,B)
=
\mathbb{E}_{b\sim p_B}
\left[
\ell(W_0+s_bA_bB_b)
\right].
\]

MatryoshkaLoRA instead uses the deterministic first-order surrogate
\[
L_{\rm multi}(A,B)
=
\sum_{r\in{\cal S}}\lambda_rf(\Delta_r)
\approx
f
\left(
\sum_{r\in{\cal S}}\lambda_r\Delta_r
\right),
\]
which becomes
\[
\ell
\left(
W_0+
A
\left[
\sum_{r\in{\cal S}}\lambda_rs_rP_r
\right]
B
\right).
\]
This gives the surrogate perturbation
$$
APB,
\qquad\text{with}\quad
P=\sum_{r\in{\cal S}}\lambda_r s_r P_r.
$$

Hence, under the smoothness and small-perturbation assumptions above, the multi-rank training problem can be approximated by optimizing a single LoRA-style objective with an intermediate diagonal weighting matrix $P$:
$$
\ell(W_0+APB).
$$

This reduction is local rather than exact, and its error is quadratic in the size of the perturbations. Still, it suggests a useful perspective: training a single pair of adapters for multiple nested ranks can be viewed as learning a shared rank-$R$ factorization whose rank components are weighted according to how strongly different truncation levels are emphasized during training.


	
\end{document}